# Aksharantar: Open Indic-language Transliteration datasets and models for the Next Billion Users

Yash Madhani[1]   Sushane Parthan[2]   Priyanka Bedekar[3]   Gokul NC[4]
Ruchi Khapra[5]   Anoop Kunchukuttan[6]   Pratyush Kumar[7]   Mitesh M. Khapra[8]

AI4Bharat[1,2,3,4,5,6,7,8]   IIT Madras[1,2,3,6,7,8]   Microsoft[6,7]

[1,2,3]{cs20s002,cs20d201,cs20m050}@smail.iitm.ac.in
[4]gokulnc@ai4bharat.org   [5]jain.ruchi03@gmail.com
[5,6]{ankunchu,pratykumar}@microsoft.com   [8]miteshk@cse.iitm.ac.in

## Abstract

Transliteration is very important in the Indian language context due to the usage of multiple scripts and the widespread use of romanized inputs. However, few training and evaluation sets are publicly available. We introduce *Aksharantar*[1], the largest publicly available transliteration dataset for Indian languages created by mining from monolingual and parallel corpora, as well as collecting data from human annotators. The dataset contains **26 million transliteration pairs** for **21 Indic languages** from **3 language families** using **12 scripts**. Aksharantar is **21 times** larger than existing datasets and is the first publicly available dataset for **7 languages** and **1 language family**. We also introduce the Aksharantar testset comprising **103k word pairs** spanning **19 languages** that enables a fine-grained analysis of transliteration models on native origin words, foreign words, frequent words, and rare words. Using the training set, we trained *IndicXlit*, a multilingual transliteration model that improves accuracy by 15% on the Dakshina test set, and establishes strong baselines on the Aksharantar testset introduced in this work. The models, mining scripts, transliteration guidelines, and datasets are available at https://github.com/AI4Bharat/IndicXlit under open-source licenses. We hope the availability of these large-scale, open resources will spur innovation for Indic language transliteration and downstream applications.

## 1 Introduction

The Indian subcontinent is home to diverse languages spanning four major language families (Indo-Aryan branch of Indo-European, Dravidian, Austro-Asiatic and Tibeto-Burman) spoken by more than a billion speakers. These languages are written in a variety of scripts: (a) Brahmi family of abugida scripts for most major Indic languages, (b) Arabic-derived abjad scripts for some languages like Urdu, Kashmiri and Sindhi, and (c) Alphabetic Roman script for many languages with recent literary history. Some of these scripts are used by multiple languages (*e.g.,* Devanagari script is used to write Hindi, Marathi, Konkani, Maithili, and Sanskrit among others; Bengali script is used to write Bengali, Assamese, and Santali).

These statistics highlight the scale and diversity of the challenge when it comes to supporting mechanisms which are convenient for typing or creating content in these diverse languages and scripts. Historically, Roman and related scripts have been widely supported across multiple platforms and device form factors for digital content creation. While native language keyboards are available in many Indic languages, most people are comfortable with the Roman keyboard. Moreover, many South Asians are multilingual and learning multiple keyboard layouts would be cumbersome. Hence, romanized input of Indian languages has become popular.

While romanized input offers a convenient solution for certain interactions, it does not solve the problem of input in the native script.

---

[1]meaning *transliteration* in Sanskrit

| Lang | Exs | Wik | Sam | Ind | Man | Tot |
|---|---|---|---|---|---|---|
| asm | - | 2 | 3 | 203 | 19 | 217 |
| ben | 104 | 107 | 193 | 1,115 | 14 | 1,337 |
| brx | - | - | - | 36 | 13 | 44 |
| guj | 111 | 8 | 67 | 1,096 | 21 | 1,236 |
| hin | 234 | 44 | 289 | 1,149 | 49 | 1,522 |
| kan | 51 | <1 | 69 | 2,930 | 27 | 3,010 |
| kas | - | <1 | - | 35 | 37 | 64 |
| kok | 65 | - | - | 619 | 37 | 702 |
| mai | 102 | 7 | - | 252 | 42 | 370 |
| mal | 61 | 1 | 59 | 4,097 | 30 | 4,195 |
| mni | - | - | - | 12 | 11 | 16 |
| mar | 60 | 26 | 49 | 1,486 | 49 | 1,594 |
| nep | - | 10 | - | 2,455 | 6 | 2,458 |
| ori | - | 1 | 23 | 380 | 13 | 398 |
| pan | 78 | 21 | 104 | 481 | 13 | 611 |
| san | - | 3 | - | 1,860 | 38 | 1,881 |
| snd | 39 | <1 | - | 53 | - | 82 |
| sin | 42 | - | - | - | - | 37 |
| tam | 71 | 1 | 61 | 3,202 | 14 | 3,301 |
| tel | 97 | <1 | 82 | 2,416 | 14 | 2,521 |
| urd | 111 | <1 | - | 649 | 3 | 748 |
| **Tot** | 1,225 | 229 | 1,000 | 24,525 | 451 | 26,345 |

Table 1: Statistics of Aksharantar dataset. All numbers are in thousands. The dataset has multiple sources (Exs: existing, Wik: Wikidata, Sam: Samanantar, Ind: IndicCorp, Man: manually collected transliterations). Tot: stands for Total unique word pairs. We use ISO 639-2 language codes throughout the article.

An optimal solution that users find beneficial is automatic transliteration of the romanized input into the native script. Hence, we undertake the creation of large-scale transliteration corpora for Indic languages along with models for transliteration of romanized inputs into native script. Note that our effort is an alternative to researching better user interface designs for native script input which can address the complexities of Indic scripts and reduce pain points for users working with multiple scripts. This should not be seen as promoting the use of Latin scripts for Indian languages, but as a practical solution to a technology gap of having good native keyboards/reduced user familiarity with native keyboards.

The following are the contributions of our work:

**Large-scale Parallel Transliteration Corpora** We build the largest publicly available parallel transliteration corpora, Aksharantar, between the Roman script and scripts for Indic languages. The corpora contains *26M* word pairs spanning *21* languages[2]. The parallel transliteration corpora has been mined from Wikidata (Vrandečić and Krötzsch, 2014), Samanantar parallel translation corpora (Ramesh et al., 2022) and IndicCorp monolingual corpora (Kakwani et al., 2020). Human judgements on a random sample of the mined corpora showed that the mined corpora is of good quality as mentioned in Section 3.5. In addition, the corpora contains a diverse set of native language words that have been transliterated manually - such data is important for input tools as they ensure coverage of words of different lengths, diverse n-grams, common as well as rare words, and named entities covering words of foreign and Indian origin. Finally, we compile existing transliteration corpora. The manually transliterated data is complemented by the mined datasets enabling us to create a large scale transliteration corpora for the purpose of building input tools. Table 1 shows the statistics of the training set.

**A diverse Transliteration Evaluation Benchmark** We create an evaluation benchmark dataset (the *Aksharantar* testset) for romanized transliteration by soliciting transliterations from native language speakers. The benchmark contains (a) Native language words with diverse n-gram characteristics, and (b) Named entities of Indic and foreign sources spanning different entity categories. This testset provides a challenging and diverse testset to benchmark transliteration performance for Indic languages. In contrast, the Dakshina testset is limited to only the most frequent words from Wikipedia. We hope this evaluation benchmark can drive progress in Indic transliteration just as diverse, multilingual benchmarks like Flores-101 (Goyal et al., 2021), WMT (Barrault et al., 2019, 2020; Akhbardeh et al., 2021), XNLI (Conneau et al., 2018) and XQuAD (Artetxe et al., 2020) have done for other NLP tasks. Table 2 shows the statistics of the benchmark set, comprised of *103K* word pairs spanning *19* languages.

---

[2]We have new data collected for 20 languages. For Sinhala, we used the Dakshina dataset data and did not collect any new data.

| Lang | Freq | Uni | NEF | NEI | Tot |
|---|---|---|---|---|---|
| **asm** | 1690 | 1938 | 742 | 1161 | 5531 |
| **ben** | 1071 | 1198 | 1059 | 1681 | 5009 |
| **brx** | 1119 | 1143 | 729 | 1145 | 4136 |
| **guj** | 2725 | 2521 | 1005 | 1517 | 7768 |
| **hin** | 1726 | 1924 | 826 | 1217 | 5693 |
| **kan** | 1851 | 2361 | 877 | 1307 | 6396 |
| **kas** | 3095 | 2588 | 816 | 1208 | 7707 |
| **kok** | 1531 | 1536 | 817 | 1209 | 5093 |
| **mai** | 1892 | 1591 | 819 | 1210 | 5512 |
| **mal** | 2261 | 2596 | 835 | 1219 | 6911 |
| **mni** | 2754 | - | 886 | 1285 | 4925 |
| **mar** | 2091 | 2375 | 831 | 1276 | 6573 |
| **nep** | 1058 | 1049 | 817 | 1209 | 4133 |
| **ori** | 1068 | 1153 | 821 | 1214 | 4256 |
| **pan** | 1049 | 1144 | 858 | 1265 | 4316 |
| **san** | 1411 | 1515 | 976 | 1432 | 5334 |
| **tam** | 1467 | 1141 | 828 | 1246 | 4682 |
| **tel** | 1105 | 1135 | 947 | 1380 | 4567 |
| **urd** | - | 2437 | 817 | 1209 | 4463 |
| **Tot** | 30964 | 31345 | 16306 | 24390 | 103005 |

Table 2: Statistics of Aksharantar testset. The testset has multiple sub-testsets (AK-Freq, AK-Uni, AK-NEF, AK-NEI). These stand for most frequent words, uniformly sampled words, foreign named entities and Indian named entities respectively.

**IndicXlit: A multilingual model for romanized to native script transliteration**
We train a multilingual model for transliteration from romanized input to native language script. Previous works and our experiments show the gains from multilingual transliteration models over monolingual models. Moreover, a single model can serve all languages making deployment and maintenance easier. Our model gives SOTA performance on the Dakshina benchmark for all the 12 intersecting languages and establishes a strong baseline on the new benchmarks released as a part of this work. Further, we show that re-ranking the top-4 transliterations with a unigram word-level language model can significantly improve transliteration accuracy for frequent words.

We make the datasets and models publicly available. The models are available under an MIT license, the Aksharantar benchmark and all data we created manually are available under the CC-BY license, whereas all the mined data is available under the CC0 license.

## 2 Related Work

**Existing Indic Language Transliteration Corpora** Very few transliteration corpora exist with Indian language-English transliterations. Table 3 summarises the statistics of existing corpora taken from various sources such as the IITB Parallel corpus (Kunchukuttan et al., 2018b), Hindi song lyrics (Gupta et al., 2012), the crowdsourced transliteration corpus (Khapra et al., 2014), the NotAI-Tech corpus (Praneeth, 2020), the BrahmiNet corpus (Kunchukuttan et al., 2015), the ILCI parallel corpus (Jha, 2010), the FIRE 2013 corpus (Roy et al., 2013), the MSR-NEWS shared task corpus (Banchs et al., 2015), the AI4Bharat-StoryWeaver corpus (Benjamin and Gokul, 2020) and the Dakshina dataset (Roark et al., 2020).

The most significant among these is the Dakshina dataset (Roark et al., 2020) which is a collection of text in both Latin and native scripts for 12 South Asian languages. It contains an aggregate of around 300k word pairs and 120k sentence pairs, with native language words sourced from Wikipedia and romanizations attested by native speaker annotators. As opposed to Aksharantar, it mostly con-sists of Indian origin words and is composed of shorter, commonly used Indic language words.

**Mining Transliteration Data** Kunchukuttan et al. (2015) mine word pairs across 10 different Indic languages from public sources such as existing parallel translation corpora and monolingual corpora. Similarly, Kunchukuttan et al. (2021) mined 600k transliteration pairs across 10 languages from publicly available parallel and monolingual sources. Compared to these existing works, we create the largest available transliteration corpora in 18 Indic languages from existing parallel translation corpora (Ramesh et al., 2022), monolingual corpora (Kakwani et al., 2020) and manual annotations from human annotators.

**Transliteration Methods** Karimi et al. (2011) compile a survey of early transliteration models, including the then state-of-the-art models grouped into generative and extractive transliteration systems. More recently, a number of transliteration systems were proposed during the Named Entities Workshop evaluation campaigns in 2018[3] (Chen et al., 2018). These campaigns comprise transliterating tasks from English to other languages with a wide variety of writing systems, including Hindi, Tamil, Bengali, Kannada, Persian, Chinese, Vietnamese, Thai and Hebrew. The transliteration models typically mentioned in the literature include a combination of neural and non-neural models. A few popular ones among these are DirecTL+ (Jiampojamarn et al., 2010), Sequitur G2P (Bisani and Ney, 2008), deep attention based RNN encoder decoder models (Kundu et al., 2018; Le and Sadat, 2018) and neural transformer based models (Merhav and Ash, 2018; Roark et al., 2020; Moran and Lignos, 2020).

**Multilingual Models** Multilingual models have been explored successfully for different NLP tasks involving Indian languages, such as, language representation modeling (Kakwani et al., 2020; Dabre et al., 2022), machine translation (Ramesh et al., 2022; Dabre et al., 2018; Goyal et al., 2020), POS tagging (Plank et al., 2016; Khemchandani et al., 2021) and named-entity recognition (Murthy and Bhattacharyya, 2016; Khemchandani et al., 2021). In the context of transliteration,

Kunchukuttan et al. (2018b) propose multilingual training for transliteration tasks, focusing on transliterations involving orthographically similar languages.

Kunchukuttan et al. (2021) also use multilingual training to train their transliteration system and recommend using single-script models to train separate models with two different language families (Indo-Aryan and Dravidian languages). As compared to that, we use a multi-script, multilingual model for all Indic languages regardless of language family group.

## 3 Mining Transliteration pairs

There are multiple sources for mining transliteration pairs using automated techniques. First, we compile publicly available transliteration corpora for existing sources. Further, we explore mining of large-scale transliteration corpora for Indian languages from parallel translation corpora, monolingual corpora and WikiData.

### 3.1 Existing sources

We gathered several existing sources. The majority of the data comes from the Dakshina corpus (Roark et al., 2020). The Dakshina corpus and the Brahminet corpus (Kunchukuttan et al., 2015) encompass multiple languages. Brahminet is mined from ILCI parallel corpus (Jha, 2010). In addition, we also compiled other small datasets, including Xlit-Crowd (Khapra et al., 2014), Xlit-IITB-Par (Kunchukuttan et al., 2018b), FIRE 2013 Track on Transliterated Search (Roy et al., 2013), NotAI-tech-English-Telugu (Praneeth, 2020), and AI4Bharat StoryWeaver Xlit Dataset (Benjamin and Gokul, 2020). Table 3 provides statistics on the compiled transliteration corpora.

### 3.2 Mining from Wikidata

Wikidata (Vrandečić and Krötzsch, 2014) is a multilingual, structured database containing items wherein an item is either an entity, a thing, a concept or term. Of interest to us, entities have *labels* which are common names of the items in multiple languages. We restrict ourselves to person and location entities since their labels will be transliterations. We extract English-Indian language label pairs cre-ating transliteration pairs. For multi-word labels, we create all possible transliteration pair candidates by taking a Cartesian product of words in English and the Indian language labels. The candidate pairs are then filtered using the automatic transliteration validator described in Section 4.4.

---
[3]NEWS 2018

|  | ben | guj | hin | kan | kok | mai | mal | mar | pan | snd | sin | tam | tel | urd |
| --- | --- | --- | --- | --- | --- | --- | --- | --- | --- | --- | --- | --- | --- | --- |
| **Dakshina** | 95 | 105 | 44 | 51 | - | - | 58 | 56 | 71 | 39 | 42 | 68 | 59 | 106 |
| **Xlit-Crowd** | - | - | 11 | - | - | - | - | - | - | - | - | - | - | - |
| **Xlit-IITB-Par** | - | - | 69 | - | - | - | - | - | - | - | - | - | - | - |
| **FIRE-2013-Track** | 5 | 1 | 36 | - | - | - | - | - | - | - | - | - | - | - |
| **AI4B-StoryWeaver** | - | - | 101 | - | 60 | 103 | - | - | - | - | - | - | - | - |
| **NotAI-tech En-Te** | - | - | - | - | - | - | - | - | - | - | - | - | 39 | - |
| **Brahminet** | 8 | 7 | 11 | - | 6 | - | 3 | 5 | 9 | - | - | 4 | 5 | 6 |
| **Total unique word pairs** | 104 | 111 | 234 | 51 | 65 | 102 | 61 | 60 | 78 | 39 | 42 | 71 | 97 | 111 |

Table 3: Statistics of transliteration pairs compiled from existing sources. All numbers are in thousands.

### 3.3 Mining from Parallel Translation Corpora

Parallel sentences can contain transliteration pairs in the form of named entities, loan words and cognates (see Table 4 for examples). To mine parallel corpora, we first learn word alignments between parallel sentences using an off-the-shelf word-aligner *viz.* *GIZA++* (Och and Ney, 2003). The aligned words can either be translations or transliterations. We use the unsupervised method suggested by Sajjad et al. (2012) (as implemented in the transliteration module (Durrani et al., 2014) of *Moses* (Koehn et al., 2007)) to mine transliteration pairs from these word alignments by distinguishing transliterations and non-transliterations. The word alignments are modeled via a linear interpolation of two generative processes: one for word transliteration and another for a non-transliteration process. The transliteration model is discovered via an iterative EM algorithm. Using this approach, we mine transliteration pairs from the *Samanantar* parallel corpora (v0.3) (Ramesh et al., 2022), the largest publicly available parallel corpora for Indian languages. The above mentioned process can result in some wrong transliteration pairs being mined. Typically, these could be leaked translation word pairs (*e.g.,* अंतर्संयुक्त→ interconnected, उपनाम→ surname, उपयोग→ Us-age, દશેરાના→ Dusshera) or highly agglutinated words on one side (અંકલેશ્વરનો → Ankleshwar).

To filter out such pairs, we use a rule-based transliteration validator which checks the correctness and coverage of consonant mappings in the word pairs. This check is sufficient for the kinds of erroneous transliteration pairs mined by the above mentioned method. The rule-based validator is described in detail in Section 4.4 in the context of the annotator interface.

### 3.4 Mining from Monolingual Corpora

Monolingual text corpora often have borrowed words from other languages (particularly English). We mine such transliteration pairs between English and Indian languages using only the list of words in the source and target languages. We use the AI4Bharat IndicCorp dataset (Kakwani et al., 2020) for the list of words for all the languages.

We first train initial multilingual transliteration models using available data (data from existing sources and mined from parallel translation corpora) in both the directions ($L_e \to L_x$, $L_x \to L_e$) and create the vocabularies of $L_e$ and $L_x$. Given word $w_x$ in $L_x$, we generate its transliteration ($w'_e$) using $L_x \to L_e$ model ($M_{xe}$). We find the new similar English words ($w_e$) from the IndicCorp corpus such that there exists at least three common 4-grams between $w'_e$ and $w_e$. The mined transliteration pair candidate ($w_x, w_e$) is scored using models in both directions.

$$s(w_x, w_e) = \frac{1}{2}\{M_{xe}(w_x, w_e) + M_{ex}(w_e, w_x)\}$$

We retain all candidate transliteration pairs with score (average log probability in both directions) greater than a threshold $t$. From our analysis of transliteration pairs across languages, we determine $t = -0.35$ as a good threshold.

| **eng** | **hin** |
|---|---|
| From the Azad Kashmir Regiment, Lt Gen Afgun has commanded a Division on the LOC when Gen Bajwa was commander of the X Corps | आजाद कश्मीर रेजिमेंट से लेफ्टिनेंट अफगुन ने एलओसी पर एक डिविजन कमांड किया है, जब जनरल बाजवा टेंथ कॉर्प्स के कमांडर थे |
| India will wear the orange jersey in match against England on June 30 in Birmingham | टीम इंडिया 30 जून को विश्व कप मैच में इंग्लैंड के खिलाफ नारंगी जर्सी में खेलेगी |
| Also read: Qualify for the Olympics, win gold at test event: All in a day's work for ace gymnast Dipa Karmakar | पढ़े: ओलिंपिक के लिए क्वालिफाई कर जिम्नास्ट दीपा कर्माकर ने रचा इतिहास |

Table 4: Examples of transliteration pairs from the Samanantar parallel translation corpus.

Some characters in low-resource languages like Oriya and Assamese are not present in existing corpora (particularly Dakshina) or corpora mined from the parallel translation corpus. For instance, the characters 'ଡ଼' in Oriya, 'ਖ਼' in Punjabi *etc.* are not found in the mined corpora. In Assamese, word pairs that dictate silent pronunciation of 'xo' character set were not present in mined corpora. Consequently, we fail to mine such pairs from the IndicCorp dataset since the transliteration models used for mining do not have these characters. Hence, we perform an additional round of mining for these low-resource languages using improved transliteration models that are trained on data gathered manually (see Section 4) that represents these missing characters.

### 3.5 Quality of the mined data

To validate the quality of the mined corpora, we perform human evaluation on a subset of mined transliteration pairs. For human evaluation, we randomly sampled 500 mined pairs equally from IndicCorp and Samanantar corpora in 12 Indic language-English pairs. Two passes of validation by different language validators were performed on this data. Validators were asked to mark the pairs which were valid transliterations. The *accuracy* of mining is defined to be the percentage of valid pairs out of the subset that was manually judged.

Table 5 shows the results of the human evaluation. We achieved minimum accuracy of 80% in each language and average accuracy of 89% across all 12 languages. Data mined from Samanantar as well as IndicCorp have high accuracy.

We analyzed the pairs judged as invalid and found that they included the following errors:

• **Vowel errors**: These include $a/e$ being added incorrectly at the end of transliterations, missing vowels, and wrong usage of vowels (*e.g.,* अमिताभ → Amtabha [missing 'i' after 'm' and unnecessary 'a' at the end]).

• **Suffix errors**: Suffixes are wrongly transliterated or missed altogether, leading to partial transliterations (*e.g.,* रोनाल्डोही→ Ronaldo, अण्णा → Anne, टोकिया→ Tokyo.

• **Named Entities**: An issue with named entities exists due to the idiosyncrasies of each individual language. A word in English might be spelled and even pronounced differently in Hindi, thereby leading to inevitable differences in its transliteration (*e.g.,* चितवन → Chhitva, गौरव→ Garhwa.

We found that most of the erroneous pairs were partial transliterations which can still be useful for training the transliteration models, introducing limited noise in the training data. The results of the human judgment and qualitative analysis confirm the high quality of the mined transliteration pairs which makes it useful for training transliteration models.

| Dataset | asm | ben | guj | hin | kan | kok | mai | mal | mar | pan | san | tam |
|---|---|---|---|---|---|---|---|---|---|---|---|---|
| **IndicCorp** | 0.91 | 0.93 | 0.91 | 0.97 | 0.98 | 0.99 | 0.91 | 0.94 | 0.97 | 0.95 | 0.78 | 0.80 |
| **Samanantar** | 0.93 | 0.92 | 0.84 | 0.76 | 0.80 | - | - | 0.80 | 0.90 | 0.86 | 0.84 | 0.80 |
| **Average** | 0.92 | 0.93 | 0.88 | 0.86 | 0.88 | 0.99 | 0.91 | 0.87 | 0.94 | 0.90 | 0.81 | 0.80 |

Table 5: Transliteration mining accuracy on a human-judged sample.

## 4 Manual Collection of Transliteration Pairs

While mining transliterations from different sources allowed us to build large transliteration corpora, this approach does not completely meet the needs of building a representative transliteration dataset for building input tools. One, an overwhelming majority of the mined corpora consists of named entities. Romanization of native words is represented only in the Dakshina dataset. Two, the Dakshina dataset only covers the most frequent words in the language as defined in Wikipedia and is good for head cases. It might not ensure diversity of native words to account for various transliteration phenomena (particularly since Wikipedia for most Indian languages is small). Three, the mined corpora only covers 12 languages for which sufficient monolingual/parallel corpora are available and which have high grapheme-to-phoneme correspondence which makes mining feasible. Four, we want to create a standard testset for transliteration in all Indic languages that is diverse and accurate.

To address these needs, we collect transliteration pairs from trained annotators for 19 Indic languages. This was a non-trivial data collection activity involving multiple annotators across India. This section describes the data collection process, quality control and logistics management. First, Indic words to be romanized are selected to ensure diversity and coverage across languages (Sections 4.1 and 4.2). Next, we collect high-quality, manually curated romanizations for these Indic words at scale by setting up a systematic process to ensure quality control and annotator productivity (Section 4.3) that is managed by a digital data collection platform (Section 4.5).

### 4.1 Sourcing Indic words

All words for manual transliteration were sourced from publicly available sources. We use the IndicCorp corpora (Kakwani et al., 2020) to source Indic language words for 11 of the 19 languages (*asm, ben, guj, hin, kan, mal, mar, ori, pan, tam, and tel*). For Maithili (*mai*), Konkani (*kok*), Bodo (*brx*), Nepali (*nep*), Kashmiri (*kas*) and Urdu (*urd*) we source words from the LDC-IL corpus (Choudhary, 2021). We collect Sanskrit (*san*) words from publicly available religious scriptures such as the Mahabharata (Sukthankar, 2017), while for Manipuri (*mni*) we source words from Wikipedia.

From the above mentioned corpora, we create unique word lists along with their frequencies. We eliminate invalid words such as those beginning with Indic language *maatras* (diacritics) (*e.g.,* ◌ँ◌ँ, ◌ँल) and words containing misplaced numerals. We further remove words of unit frequency from our word list. Table 6 shows the word-list size for each language from where the final words for transliteration are selected.

### 4.2 Selecting a diverse set of words for manual transliteration

We select native script words for manual transliteration with the goal of ensuring coverage of words of different lengths, coverage of diverse n-grams, common as well as rare words, and foreign origin words. While selecting the source words, we ensure that these words are not already covered in the sources mentioned in Section 3. We use a combination of the following methods for selecting words for transliteration:

• **Most frequent words**: To account for the most frequent words in a language, we select the top 5000 words for each language. Specifically, we would like to point that the sampled words are not already present in the Dakshina dataset (Roark et al., 2020) and in fact supplement them, thereby augmenting the collection of most frequent words.

| Language | asm | ben | guj | hin | kan | kok | mai | mal | mar | ori | pan | san | tam | tel | nep | brx | kas | urd | mni |
|---|---|---|---|---|---|---|---|---|---|---|---|---|---|---|---|---|---|---|---|
| Count | 292 | 1867 | 1903 | 1456 | 3926 | 314 | 264 | 5895 | 2164 | 597 | 876 | 48 | 3874 | 3147 | 501 | 254 | 94 | 83 | 15 |

Table 6: Total unique word-list extracted from publicly available corpora from which source words for transliteration were sampled. All numbers are in thousands.

• **N-gram Diversity** The probabilities from a character language model would be a good indicator of the frequency of n-grams in a word. Hence, we train a 4-gram character LM over all words for each language using KenLM with Kneser-Ney smoothing (Heafield, 2011; Heafield et al., 2013). We compute log probability scores (normalized by word length and scaled to 0-1 range) for each candidate word using the character LM. The words are then sharded into 10 bins corresponding to the 10 probability deciles. Each bin would represent different character n-gram phenomena. Hence, words are uniformly sampled from each bin thus ensuring n-gram diversity in the source words. By obtaining transliterations for words with diverse n-grams, we complement mined corpora which are mostly composed of named entities and head inputs. We sampled a total of 10000 words using this method for each language.

• **Named Entities** Named entities are well-represented in the data mined from different sources as mentioned in Section 3. The purpose of manual collection of named entity transliterations was to create a test set of named entities from diverse categories in each language. We sampled 2000 named entities in English spanning 3 broad categories: Names, locations, and organisations. These cover both Indian origin and foreign origin words. We sourced names (both Indian and foreign personal names) as well as locations by randomly sampling words from collections on websites dedicated for the same. Organisation names are sourced from the stock market library list of 1600+ companies listed in NSE[4]. 2000 named entities in total were divided into 800 names (400 each of Indian and foreign origin), 800 locations (400 each of Indian and foreign origin) and 400 Indian organisations.

[4]Stock market library

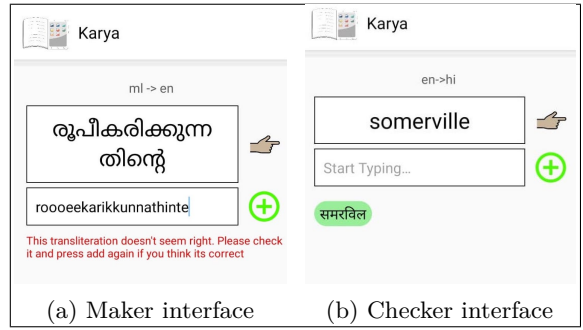

(a) Maker interface  (b) Checker interface

Figure 1: Annotation UI in the *Karya* app.

### 4.3 Annotation Process and Quality Control

We collect transliterations via a two-step process akin to a maker-checker process. The *transliterator* creates multiple romanized variants for a native word and the correctness of the transliterations is checked by a *validator*, who also has the freedom to enter word variants if they see fit. Transliterators use a mobile application with the user interface shown in Figure 1a to enter the transliterations. We conducted multiple pilot projects to study different annotation styles, identify common annotation errors made, and incorporated a set of constraints and instructions. While annotators are free to enter all common variants, they were encouraged to follow the basic instructions as much as possible.

• Through the pilot projects, it was observed that setting a high maximum number of variants (such as 10) led to annotators attempting all possible transliterations (including erroneous ones) for a given Indic word. Therefore, the maximum number of variants is capped to 4 per transliterator and 2 per validator.

• Transliterators are instructed to avoid frivolous variants, like duplications of vowels, unless they are important.

• A rule-based automatic transliteration validator is provided to flag potentially wrong transliterations. The transliterator can choose

to ignore the transliteration validator as per their discretion.

The validator can reject wrong variants as well as enter any important variants for a native script word missed by the transliterator on a mobile app using the interface shown in Figure 1b. The variants accepted or added by the validator constitute the final set of romanized variants for the input word.

### 4.4 Automatic Transliteration Validator

To aid the transliterators, we provide an automatic rule-based transliteration checker. The checker flags potentially wrong transliterations - helping the transliterator correct any mistake in entering the romanized characters. Typically, we found that the rule-based transliteration validator helped identify typographical errors and other mistypings. Such a checker helps the transliterator perform the task well, ensures consistency in transliterations and preserves wastage of validator effort. Note that the automatic checker is only a guide to the transliterators, who can override its checks at their discretion.

The transliteration checker is based on the Transliteration Equivalence algorithm for English (Roman)-Hindi described in Khapra et al. (2014) which basically checks equivalence of the consonant mappings in a potential transliteration pair. To achieve this, the algorithm takes two pieces of information: (i) a stop-list of vowels in the two languages, and (ii) a list of consonant mappings between the two languages. We extend the above mentioned approach to all the 14 languages by incorporating the above mentioned rules for each language with the aid of language experts. For instance, Table 7 shows the consonant mapping for Kannada language. There is a large overlap in the consonant mapping rules across languages, but the checker incorporates language-specific exceptions as well. The transliteration validator firstly removes all characters which are either vowels or present in the stop-list, from the English variant. The checker then sequentially maps each English consonant to the relevant Indic language consonant according to the language mapping table as shown in Table 7. Once all possible Indic language variants of the English word are formulated by the checker, it compares them against the original Indic word to check validity of the romanized transliteration. We checked the effectiveness of the automatic transliteration validator on transliteration pairs in the Dakshina train set. It had a minimum accuracy of 90% for most languages as shown in Table 8, indicating its utility and non-intrusiveness.

### 4.5 Annotation Platform

Building a diverse transliteration dataset is a complex process involving liaising with numerous annotators working remotely from different parts of India across multiple data annotation agencies. To manage data collection and annotators as well as ensure quality control we use *Project Karya* (Chopra et al., 2019; Abraham et al., 2020), an open source crowdsourcing platform developed by Microsoft Research, which harnesses the current trend of cheap, accessible smartphones; and employ this to make digital language work more inclusive and accessible to the local population. The app was not open to local crowd. We use Karya for collecting transliteration data from annotators chosen from the pilot tasks that were conducted by us. Each transliteration micro-task contains 100 native words to be transliterated and then validated post transliteration. The interface is as shown in Figure 1.

### 4.6 Annotator Information

We employ 68 annotators from two data annotation agencies as transliterators and validators. The annotators are native language speakers and are proficient in English. Validators were annotators with more experience in linguistic tasks. We ran some pilot tasks with annotators from the agencies and for the larger task we shortlisted annotators making less than 5% errors on pilot tasks. The annotators were paid INR 2 (USD 0.026) per native language word.

## 5 The Aksharantar Dataset

We consolidate the mined dataset (Section 3) and manually collected dataset (Section 4) and then create train, validation and test splits for the Aksharantar dataset. Table 9 shows the statistics of the train and validation splits.

| | | | | | | | | | | | |
|---|---|---|---|---|---|---|---|---|---|---|---|
| b | ಬ ಭ | | l | ಲ ಳ | | t | ಟ ಱ ತ ಥ ಶ ಷ ಚ ದ | | | | |
| c | ಕ ಚ ಛ ಶ ಷ ಸ | | m | ಮ | | v | ವ | | | | |
| d | ಡ ಢ ದ ಧ | | n | ಣ ನ | | w | ವ | | | | |
| f | ಫ಼ | | p | ಪ ಫ | | x | ಕಸ | | | | |
| g | ಗ ಘ಼ ಜ಼ ಜ಼ | | q | ಕ಼ | | z | ಜ಼ ಱು | | | | |
| j | ಜ ಱು | | r | ಖೂ ರ ಖು | | | | | | | |
| k | ಕ ಖ | | s | ಶ ಷ ಸ ಜ಼ ಱು | | | | | | | |

Table 7: Kannada consonant mapping table.

| ben | guj | hin | kan | mal | mar | pan | tam | tel |
|---|---|---|---|---|---|---|---|---|
| 0.90 | 0.97 | 0.95 | 0.98 | 0.84 | 0.98 | 0.97 | 0.93 | 0.96 |

Table 8: Accuracy of Automatic Transliteration Validator on Dakshina dataset.

The testset is created purely from the manually collected dataset. The following partitions are defined in the testset:

- **AK-Freq**: contains source words selected by word frequency.
- **AK-Uni**: contains source words selected by uniform sampling described earlier.
- **AK-NEF**: contains foreign-origin named entities.
- **AK-NEI**: contains Indian-origin named entities.

These sub-testsets help to evaluate performance of transliteration models on specific categories of words. Table 2 shows the statistics of the testset.

While creating the testset, we strictly ensure that there is no word-overlap between any training and test/validation sets for inference. Note that the testsets considered for overlap computation include the Dakshina testset. A transliteration pair *(en, t)* will be removed from the training set if (i) the Latin script word *en* is present in the romanised validation/test set of any language pair, or (ii) the Indic script word *t* is present in the Indic language validation/test set of any language pair. Being a jointly trained model, it is necessary to ensure an *en* word in the test/validation set of one language is not part of the training set of any other language pair. For example, an *en* word in the *en-ta* validation/test set cannot be part of a word-pair present in the training set of any other language.

## 6 IndicXlit: A Multilingual Model for Transliteration

With the parallel transliteration corpora described in Sections 3 and 4, we train a transliteration model *viz.* IndicXlit for transliterating romanized Indic language input to native script. IndicXlit is a single multilingual, multiscript transliteration model that supports 21 Indic languages. We train a joint model mode since: (a) low-resource languages would benefit from transfer learning, (b) previous works show that multilingual transiteration models are better at generating canonical spellings (Kunchukuttan et al., 2018a), and (c) deployment and maintenance are easier since only a single model has to be supported. In this section, we describe the model architecture and training details for IndicXlit.

**Model Architecture** We use a transformer based encoder-decoder architecture (Vaswani et al., 2017). It is a multilingual character level transliteration model (Kunchukuttan et al., 2021) in a one-to-many setting *i.e.,* the model consumes a romanized character sequence (Roman script) and generates an output character sequence in the Indic language script. The input sequence includes a special *target language tag* token to specify the target language (John-son et al., 2017). The input vocabulary is the set of Roman characters found in the training set, while the output vocabulary is the union of characters from various Indic language scripts found in the training set. The input and output vocabulary sizes are 28 and 780 characters respectively.

| Split | asm | ben | brx | guj | hin | kan | kas | kok | mai | mal | mni | mar | nep | ori | pan | san | snd | sin | tam | tel | urd | Total |
|---|---|---|---|---|---|---|---|---|---|---|---|---|---|---|---|---|---|---|---|---|---|---|
| Training | 179 | 1,231 | 36 | 1,143 | 1,299 | 2,907 | 47 | 613 | 283 | 4,101 | 10 | 1,453 | 2,397 | 346 | 515 | 1,813 | 60 | 32 | 3,231 | 2,430 | 699 | 24,823 |
| Validation | 4 | 11 | 3 | 12 | 6 | 7 | 4 | 4 | 4 | 8 | 3 | 8 | 3 | 3 | 9 | 3 | 8 | 4 | 9 | 8 | 12 | 133 |

Table 9: Training and validation set statistics for Aksharantar. All numbers are in thousands.

We use Fairseq (Ott et al., 2019) for training our transliteration models, specifically the translation multi simple epoch task. The model has 6 encoder and 6 decoder layers, 256 dimensional input embeddings, feedforward network (FFN) dimension of 1024 and 4 attention heads. We use the GELU activation function (Hendrycks and Gimpel, 2016) in the feedforward layer, and dropout value of 0.5. We preprocess multi-head attention, encoder attention and each layer of FFN with layernorm. We also add layer normalization to the embeddings (Ba et al., 2016). The model size is **11** million parameters.

**Training Details** We optimize the cross-entropy loss using the Adam optimizer (Kingma and Ba, 2015) with Adam-betas of (0.9, 0.98). We use a peak learning rate of 0.001, 4000 warmup steps and the *inverse-sqrt* learning rate scheduler. We use a global batch size of 4096 pairs. Each minibatch contains examples from all language pairs. Due to the skew in data distrubution across languages, we use temperature sampling (Arivazhagan et al., 2019) to oversample data from low-resource languages with temperature $T = 1.5$. We optimize the above mentioned values of the hyperparameters over the Dakshina training and development set. Table 10 describes the hyperparameters valued we experimented with while tuning. We train the model on 4 A100 GPUs for a maximum of 50 epochs.

**Decoding** We use beam search with beam size = 4. In addition, we also rescore top-4 candidates using a revised score $F_c$ generated by interpolating a word-level unigram LM score ($P_c$) and transliteration score ($T_c$) as shown

| Hyperparameters | Values |
|---|---|
| embed-dim | {128, 256, 512} |
| ffn-embed-dim | {1024, 2048} |
| layers | {4, 6, 8} |
| dropout | {0.2, 0.36, 0.5} |
| learning-rate | 0.001 to 0.0001 |
| warmup-updates | 2000 to 12000 |

Table 10: Hyperparameters used for tuning with values.

below.

$$F_c = \alpha T_c + (1 - \alpha) P_c$$

We use $\alpha = 0.9$ based on tuning the parameter on the development set.

## 7 Analysis of IndicXlit transliteration quality

In this section, we analyze the transliteration quality of IndicXlit on various testsets. The Dakshina testset is an existing, publicly available testset, while the Aksharantar testset is a diverse testset introduced in this work.

### 7.1 Performance on Dakshina testset

We compare the accuracy of IndicXlit with the best reported results on the Dakshina testset (in Table 11). Note that the Dakshina testset covers only 12 of the languages that are part of the Aksharantar dataset. We observe that the IndicXlit model significantly improves the results reported by (Roark et al., 2020) on the Dakshina dataset, with a 15% improvement in average accuracy across languages. Since the size of training data is a major difference between the two models, it is clear that large-scale mined transliteration pairs help to significantly improve the transliteration quality. In addition, multilingual training also helps improve the transliteration quality. This can be seen from the bilingual and multilingual models we trained on the Dakshina training set. These observations are further supported by ablation results reported in Section 8. The largest improvements are seen for *mar* (30.3%) and *guj* (25.7%), possibly because they are similar to the high resource hin language and *mar* also shares the script with Hindi. The least improvements are seen for *tam* (4.6%) and *tel* (8.9%).

| Model | ben | guj | hin | kan | mal | mar | pan | snd | sin | tam | tel | urd | avg |
|---|---|---|---|---|---|---|---|---|---|---|---|---|---|
| Roark et al. (2020) | 49.40 | 49.50 | 50.00 | 66.20 | 58.30 | 49.70 | 40.90 | 33.20 | 54.70 | 65.70 | 67.60 | 36.70 | 51.83 |
| *Our models trained on Dakshina dataset* | | | | | | | | | | | | | |
| Bilingual | 41.85 | 42.79 | 46.74 | 58.35 | 52.86 | 41.47 | 37.37 | 35.09 | 52.41 | 56.04 | 63.27 | 34.74 | 46.91 |
| Multilingual | 47.20 | 51.04 | 51.80 | 66.45 | 56.59 | 51.05 | 42.27 | 41.37 | 58.77 | 63.56 | 67.13 | 38.38 | 52.97 |
| **IndicXlit** | **55.49** | **62.02** | **60.56** | **77.18** | **63.56** | **64.85** | **47.24** | **48.56** | **63.91** | **68.10** | **73.38** | **42.12** | **60.58** |

Table 11: Comparing Top-1 accuracies reported on the Dakshina test set.

| Testset | asm | ben | brx | guj | hin | kan | kas | kok | mai | mal | mni | mar | nep | ori | pan | san | tam | tel | urd | avg |
|---|---|---|---|---|---|---|---|---|---|---|---|---|---|---|---|---|---|---|---|---|
| Dakshina | - | 55.49 | - | 62.02 | 60.56 | 77.18 | - | - | - | 63.56 | - | 64.85 | - | - | 47.24 | - | 68.10 | 73.38 | 42.12 | 61.45 |
| AK-Freq | 65.95 | 63.03 | 74.80 | 65.36 | 58.61 | 80.69 | 31.24 | 65.38 | 78.65 | 71.67 | 83.19 | 74.69 | 80.17 | 66.79 | 49.00 | 81.56 | 73.76 | 90.05 | - | 69.70 |
| AK-Uni | 55.10 | 60.47 | 66.75 | 58.17 | 52.99 | 72.65 | 27.88 | 61.16 | 64.30 | 58.68 | - | 54.00 | 79.94 | 51.95 | 32.13 | 75.92 | 64.62 | 79.31 | 48.38 | 59.13 |
| AK-NEF | 38.90 | 36.43 | 30.82 | 45.60 | 55.87 | 53.25 | 13.22 | 27.26 | 33.37 | 29.45 | 44.62 | 49.51 | 49.14 | 29.62 | 31.17 | 19.58 | 39.00 | 53.55 | 48.04 | 38.34 |
| AK-NEI | 39.16 | 40.50 | 30.89 | 51.53 | 61.41 | 48.72 | 25.06 | 39.50 | 49.50 | 37.81 | 44.63 | 56.61 | 55.45 | 32.15 | 40.12 | 26.76 | 44.63 | 51.57 | 47.69 | 43.35 |
| Micro-avg | 52.82 | 54.06 | 52.39 | 60.46 | 58.30 | 72.06 | 26.17 | 51.70 | 61.23 | 59.29 | 66.84 | 62.55 | 66.72 | 45.48 | 43.88 | 56.43 | 63.99 | 71.74 | 43.83 | 56.31 |

Table 12: Top-1 accuracy for IndicXlit on various testsets.

### 7.2 Performance on Aksharantar testset

We report the accuracy of IndicXlit on the Aksharantar testset (in Table 12), particularly looking at the accuracy on various sub-testsets to understand model performance on different categories of words. The following are the major observations:

*Frequent words are easier.* The performance on Dakshina dataset and the AK-Freq dataset, both comprised of frequent words in the language, is similar. The AK-Freq testset has the best performance across all subtestsets, suggesting that this test set is easiest to transliterate. These words are shorter on average and might also be comprised of common n-grams - explaining the good performance.

*Words with diverse n-grams are harder.* On the other hand, the AK-Uni testset comprised of uniformly sampled words with diverse n-gram characteristics, is much more challenging with average accuracy being 10 points lower than the AK-Freq testset. This testset presents a challenging usecase for transliteration systems. The lower accuracy on this testset can be attributed to the average length of words and rarity of the n-grams.

*Named entities are the hardest.* The named entity testsets are the most difficult testsets, particularly foreign named entities, even though named entities constitute a large fraction of the mined training data. The performance of foreign named entities is not surprising since the grapheme-phoneme mismatch is larger for these entities. While Indian named entities perform better than foreign named entities, their transliteration accuracy is still lower than the uniformly sampled testset. This is surprising and warrants further investigation.

*Some languages are harder.* In terms of language-wise accuracy, the lowest-performing languages are ones using the Arabic script (*urd, kas*) or those with lesser training data (*asm, brx, ori*).

*Re-ranking helps on average.* Unigram re-ranking of the candidates helps improve the transliteration accuracy significantly by 12% on an average across languages (See Table 13 for results). LM re-ranking mostly benefits the native language words and high resource languages with a lot of monolingual data for training LMs.

*Re-ranking doesn't help for named entities.* Unigram re-ranking shows limited benefits for named entities. This is not surprising since named entities might not be well represented in the LM given their rarity. Similarly, low-resource languages with limited monolingual data benefit less from LM re-ranking. Rare words thus pose a challenge to the quality of transliteration models.

## 7.3 Error analysis

We performed a manual analysis of the IndicXlit outputs to understand the errors in model output. For this analysis, we randomly sample 100 words each for Bengali, Gujarati, Hindi, Kannada, Marathi, Punjabi and Telugu from the Dakshina dataset. Table 14 summarizes the major transliteration errors as described below.

*Vowels.* The most common errors across languages are with respect to vowels, as reported in previous studies (Kunchukuttan et al., 2021). Insertion/deletion of the 'ा' vowel diacritic along with confusion between short/long vowel diacritics constitute a large fraction of transliteration errors.

*Similar consonants.* Another common source of errors is confusion between similar consonants as shown in Table 14.

*Gemination.* Other prominent errors are with respect to gemination (*e.g.,* {input: *thathvavethaga*, reference: తత్వ్వేథగా, prediction: తత్వ్వేథగా}, {input: *vittannanni*, reference: వి-త్తన్నాన్ని, prediction: విత్తన్నాన్ని}).

*Acronyms.* Acronyms have a peculiar transliteration behaviour which needs to be handled differently (*e.g.,* {input: *wsd*, reference: ಡಬ್ಲ್ಯೂ-ಎಸಡಿ, prediction: ವಾಸ್ಡ}, {input: *spwd*, reference: एसपीडब्ल्यूडी, prediction: स्प्वड}).

*Contextual ambiguity.* The "other errors" category is the result of ambiguities which cannot be easily resolved from character context alone. These are prevalent across all the testsets to varying degrees.

*Language specific.* In addition, we observed some language specific error categories. For example, in Gujarati, there is ambiguity between 'ં' and 'ન' characters; in Marathi, there are instances of deletion of 'ं' diacritic; in Punjabi, there are instances of addition/deletion of 'ਥ', 'ੰ' and 'ੱ' vowels/diacritics; in Bengali, there is ambiguity between 'শ' and 'স'; in Kannada, confusion exists between consonants 'ಥ' and 'ಸ', as well as 'ಳ' and 'ಲ'; similarly in Telugu, there is ambiguity between 'ఘ' and 'ల'.

*Valid alternatives.* Finally, some of the reported transliteration errors are actually valid alternative transliterations (*e.g.,* {input: *khurasan*, reference: खुरासान, prediction: खु-रासन}, {input: *bayern*, reference: बायर्न, prediction: बेयर्न}).

## 8 Ablation Studies

In this section, we describe various ablation studies and their results which drove the design choices of the IndicXlit model described in Section 6. We discuss the research questions investigated in this ablation study and their results. The results are summarized in Table 15. The ablation studies were carried out on the Dakshina testset for 9 languages *viz. ben, guj, hin, kan, mal, mar, pan, tam, tel*. The following are the various research questions we studied:

**Impact of various transliteration corpora sources.** We investigate if the addition of various transliteration corpora we collected improve transliteration accuracy over the baseline results on the Dakshina training set. To this end, we train separate monolingual models for each language. We initially trained a baseline model by using just the Dakshina training set, followed by successive addition of transliteration pairs collected/mined from various sources. We observe a consistent increase in transliteration quality as transliteration pairs from various sources are added. Particularly, we observe a significant improvement in performance when we add the word pairs mined from the monolingual corpora, IndicCorp, which constitutes the largest component of Aksharantar. Further, addition of the manually collected transliteration pairs described in Section 4 does not have an impact on these languages and the Dakshina testset since IndicCorp already contains sufficient data to model the frequent words that are part of the Dakshina testset. However, as shown in Table 16, we do observe that the manually collected data improves the micro-averaged transliteration accuracy over Dakshina and all Aksharantar testsets *viz.* AK-Freq, AK-Uni, AK-NEF, AK-NEI. This suggests that the manually collected data improves accuracy on other testset categories. Moreover, the manual data is necessary for extremely low-resource languages where there is no data in the public domain and for bootstrapping mining efforts.

**Impact of Multilingual Models.** How do multilingual models compare with monolin-gual models across languages? We see that multilingual models show a slight improvement over the monolingual results on the Dakshina benchmark. In another experiment, we compare monolingual and multilingual models (for *18* languages) using all sources (except the manually collected datasets). In this case, we see significant increase in accuracy for low-resource languages using multilingual models (Table 17). Thus multilingual models significantly improve performance for low-resource languages, while at least retaining performance on high-resource languages with a single model.

| Testset | asm | ben | brx | guj | hin | kan | kas | kok | mai | mal | mni | mar | nep | ori | pan | san | tam | tel | urd | avg |
|---|---|---|---|---|---|---|---|---|---|---|---|---|---|---|---|---|---|---|---|---|
| **Dakshina** | - | 55.49 | - | 62.02 | 60.56 | 77.18 | - | - | - | 63.56 | - | 64.85 | - | - | 47.24 | - | 68.10 | 73.38 | 42.12 | 61.45 |
| +rerank | - | 69.41 | - | 73.84 | 72.44 | 85.28 | - | - | - | 73.59 | - | 76.18 | - | - | 60.45 | - | 78.53 | 84.46 | 46.98 | 72.12 |
| **AK-Freq** | 65.95 | 63.03 | 74.80 | 65.36 | 58.61 | 80.69 | 31.24 | 65.38 | 78.65 | 71.67 | 83.19 | 74.69 | 80.17 | 66.79 | 49.00 | 81.56 | 73.76 | 90.05 | - | 69.70 |
| +rerank | 77.44 | 79.74 | 78.42 | 84.57 | 67.94 | 90.54 | 30.04 | 76.29 | 87.57 | 83.36 | 91.76 | 85.47 | 86.62 | 79.28 | 60.46 | 90.07 | 85.89 | 94.76 | - | 79.46 |
| **AK-Uni** | 55.10 | 60.47 | 66.75 | 58.17 | 52.99 | 72.65 | 27.88 | 61.16 | 64.30 | 58.68 | - | 54.00 | 79.94 | 51.95 | 32.13 | 75.92 | 64.62 | 79.31 | 48.38 | 59.13 |
| +rerank | 67.20 | 69.22 | 65.00 | 68.72 | 63.08 | 82.18 | 27.12 | 58.95 | 62.79 | 69.43 | - | 63.30 | 82.71 | 63.51 | 42.67 | 88.06 | 76.53 | 86.09 | 46.11 | 65.70 |
| **AK-NEF** | 38.90 | 36.43 | 30.82 | 45.60 | 55.87 | 53.25 | 13.22 | 27.26 | 33.37 | 29.45 | 44.62 | 49.51 | 49.14 | 29.62 | 31.17 | 19.58 | 39.00 | 53.55 | 48.04 | 38.34 |
| +rerank | 37.01 | 36.31 | 28.90 | 47.43 | 59.17 | 56.44 | 13.10 | 30.07 | 35.82 | 30.55 | 42.91 | 51.71 | 55.62 | 28.40 | 34.11 | 18.60 | 42.91 | 55.99 | 51.83 | 39.84 |
| **AK-NEI** | 39.16 | 40.50 | 30.89 | 51.53 | 61.41 | 48.72 | 25.06 | 39.50 | 49.50 | 37.81 | 44.63 | 56.61 | 55.45 | 32.15 | 40.12 | 26.76 | 44.63 | 51.57 | 47.69 | 43.35 |
| +rerank | 41.31 | 43.14 | 29.49 | 54.18 | 67.93 | 52.20 | 28.87 | 42.56 | 55.79 | 41.13 | 44.79 | 61.82 | 62.40 | 33.22 | 42.76 | 29.33 | 47.69 | 55.29 | 52.73 | 46.66 |
| **Micro-avg** | 52.82 | 54.06 | 52.39 | 60.46 | 58.30 | 72.06 | 26.17 | 51.70 | 61.23 | 59.29 | 66.84 | 62.55 | 66.72 | 45.48 | 43.88 | 56.43 | 63.99 | 71.74 | 43.83 | 56.31 |
| +rerank | 60.78 | 65.90 | 52.04 | 72.19 | 68.14 | 79.97 | 26.20 | 55.47 | 65.57 | 68.55 | 71.51 | 72.17 | 72.32 | 51.81 | 54.72 | 62.99 | 73.53 | 80.22 | 47.49 | 63.24 |

Table 13: Top-1 accuracy by re-ranking top 4 candidates for IndicXlit on various testsets.

| Types of errors | % | Most common errors across all languages |
|---|---|---|
| Vowel errors | 45 | Vowels are getting interchanged, model is skipping or adding 'ा' |
| Interchanging short, long vowels | 15 | {ि ⇌ ी}, {ि ⇌ ी}, {ु ⇌ ू}, {ो ⇌ ो}, {ो ⇌ ो}, {े ⇌ ै}, {आ ⇌ अ} |
| Consonant errors | 25 | {ड ⇌ द}, {त ⇌ ट}, {ण ⇌ न} |
| Other errors | 15 | Acronyms, gemination errors, silent characters, *valid* alternative transliterations, unnecessary vowel suppressor addition |

Table 14: Summary of error analysis of IndicXlit outputs.

| No | Description | ben | guj | hin | kan | mal | mar | pan | tam | tel | avg |
|---|---|---|---|---|---|---|---|---|---|---|---|
| *Impact of various transliteration sources (bilingual models)* | | | | | | | | | | | |
| (1) | Dakshina baseline | 41.85 | 42.79 | 46.74 | 58.35 | 52.86 | 41.47 | 37.37 | 56.04 | 63.27 | 48.97 |
| (2) | (1)+Existing | 41.92 | 43.08 | 48.67 | 58.91 | 51.44 | 43.48 | 38.75 | 58.58 | 65.19 | 50.00 |
| (3) | (2)+Wikidata | 44.24 | 43.90 | 49.08 | 57.75 | 50.39 | 45.81 | 40.07 | 57.16 | 63.83 | 50.25 |
| (4) | (3)+Samanantar | 48.47 | 47.48 | 53.11 | 64.15 | 55.68 | 49.02 | 40.19 | 62.14 | 67.76 | 54.22 |
| (5) | (4)+IndicCorp | 56.00 | 60.09 | 56.33 | 76.30 | 64.82 | 65.40 | 46.05 | 67.72 | 73.37 | 62.90 |
| (6) | (5)+Manual | 56.07 | 59.15 | 58.44 | 76.82 | 62.71 | 64.69 | 45.44 | 65.78 | 74.14 | 62.58 |
| *Impact of multilinguality and script unification* (baseline: (5)) | | | | | | | | | | | |
| (7) | Multi-script | 54.94 | 60.89 | 58.89 | 76.72 | 64.05 | 64.25 | 47.66 | 67.45 | 73.12 | 63.11 |
| (8) | Single-script | 55.42 | 61.92 | 58.26 | 77.52 | 64.88 | 65.20 | 47.31 | 68.23 | 73.40 | 63.57 |
| *Impact of language family specific models* (baseline: (7)) | | | | | | | | | | | |
| (9) | IA languages *(57.33)* | 56.77 | 61.92 | 59.59 | | | 65.56 | 48.20 | | | 58.41 |
| (10) | DR languages *(70.34)* | | | | 77.52 | 64.61 | | | 68.64 | 73.84 | 71.15 |
| *Impact of re-ranking with unigram LM* (baseline: (7)) | | | | | | | | | | | |
| (11) | $\alpha = 0.9, k = 4$ | 69.00 | 72.24 | 71.45 | 85.64 | 74.53 | 75.47 | 59.68 | 77.83 | 83.85 | 74.41 |
| (12) | $\alpha = 0.9, k = 10$ | 74.03 | 75.71 | 73.05 | 87.53 | 77.98 | 79.76 | 61.82 | 81.66 | 86.69 | 77.58 |
| (13) | $\alpha = 0.8, k = 4$ | 71.13 | 74.33 | 72.02 | 87.05 | 75.61 | 77.36 | 61.02 | 79.69 | 85.35 | 75.95 |
| (14) | $\alpha = 0.8, k = 10$ | 70.77 | 73.14 | 68.69 | 86.38 | 75.28 | 78.10 | 58.62 | 79.51 | 83.60 | 74.90 |

Table 15: Top-1 accuracies from experiments in the ablation study. For experiments (9) and (10) the average accuracy for the baseline pan-Indic model (7) are indicated in red in column 2.

**Impact of script unification.** How do multi-script models compare with single script models? Scripts of Indic languages originate from the ancient Brahmi script. Although each Indic script has a unique Unicode codepoint range, a 1-1 mapping between most characters of different scripts is possible since the Unicode standard accounts for similarities between Indic scripts. This can potentially improve transfer learning between languages. We experiment with *single script models* converting characters from all Brahmi-derived scripts to the Devanagari script using the IndicNLP library (Kunchukuttan, 2020). A special language token is added to every input sequence to distinguish the original Indic language, as described in Section 6. After decoding, the Devanagari script output is converted back to the target language's Indic script using the 1-1 mapping. We observe that single-script and multi-script models have similar performance. Given the small difference and negligible model size overhead, we opt to use a multi-script model for all Indic languages to simplify pre-processing of data and incorporation of scripts such as the Arabic script, which cannot be easily mapped to the Devanagari script.

**Impact of language family specific models.** Are language-family specific models better than a single model for all Indic languages? We observe that language-family specific models are slightly better than a pan-Indic model. Given the small difference in quality and the convenience of maintaining and deploying a single model, we choose to train IndicXlit as a pan-Indic language model.

**Impact of re-ranking candidate transliterations.** Can transliteration accuracy be improved by re-ranking the top-k transliteration candidates using a word-level language model? We train a unigram word-level language model and rescore the output as described in Section 6. We observe a 12% and 14% improvement in accuracy by rescoring the top-4 and top-10 candidates respectively with appropriate $\alpha$. For the IndicXlit model, we finally re-score the top-4 candidates using $\alpha = 0.9$.

## 9 Conclusion and Future work

In this work, we take a major step towards creating publicly available, open datasets and open-source models for transliteration in Indic languages. We introduce Aksharantar, the largest transliteration parallel corpora for 21 Indic languages containing 26 million transliteration pairs, and covering 20 of the 22 languages listed in the Indian constitution. This corpus was collected via a combination of large-scale transliteration mining, manual collection of diverse transliterations and intermediate model building creating a positive feedback loop. While large-scale mining helps create a large dataset in an inexpensive way, manual collection ensures diversity of words and good-coverage for low-resource languages. We also create a diverse, high-quality testset for romanized to Indic script transliteration covering words pairs with various characteristics and enabling fine-grained analysis of different transliteration usecases. We also build IndicXlit, a transformer-based transliteration model, for romanized input to Indic script transliteration. IndicXlit achieves state-of-the-art results on the Dakshina testset. We also provide baseline results on the new Aksharantar testset along with a qualitative analysis of the model performance.

The dataset and models will be available publicly under a permissive license. We hope the dataset will spur innovations in transliteration and its downstream applications in the Indian NLP space. In the future, we plan to release transliteration models for Indic to Roman script transliteration.

| Dataset | ben | guj | hin | kan | mal | mar | pan | tam | tel | avg |
|---|---|---|---|---|---|---|---|---|---|---|
| All | **54.15** | **58.54** | **56.68** | **71.98** | 57.91 | 59.90 | 41.94 | 61.14 | **72.08** | **59.37** |
| No manual | 50.99 | 38.39 | 54.56 | 71.16 | **58.91** | **60.29** | **42.91** | **62.95** | 71.97 | 56.90 |

Table 16: Impact of manually collected pairs (micro-averaged accuracy over all testsets).

| Model | asm | ben | guj | hin | kan | kok | mai | mal | mar | nep | ori | pan | tam | tel | urd | avg |
|---|---|---|---|---|---|---|---|---|---|---|---|---|---|---|---|---|
| Monolingual | 24.74 | 50.99 | 56.21 | 54.56 | 71.16 | 38.39 | 36.78 | **58.91** | **60.29** | 14.78 | 24.08 | 42.91 | 62.95 | **71.97** | 31.87 | 46.71 |
| Multilingual | **29.32** | **51.68** | **57.24** | **56.07** | **71.60** | **44.98** | **52.75** | 58.90 | 60.25 | **44.20** | **27.75** | **43.94** | **63.41** | 71.27 | **38.84** | **51.48** |

Table 17: Monolingual *vs.* multilingual models (micro-averaged accuracy over all testsets).

## Limitations

The benchmark for transliteration for the most part contains clean words (grammatically correct, single script, etc.). Data from the real world might be noisy (ungrammatical, mixed scripts, code-mixed, invalid characters, etc.). A better representative benchmark might be useful for such use cases. However, the use-cases captured by this benchmark should suffice for the collection of clean transliteration corpora. This also represents a first step for many low-resource languages where no transliteration benchmark exists.

In this work, training data is limited to the 20 languages and test data is limited to 19 languages listed in the $8^{th}$ schedule of the Indian constitution. Further work is needed to extend the benchmark to many more widely used languages in India (which has about 30 languages with more than a million speakers). **Subsequent to the acceptance of this work, we have also released training and testsets for one more Indic language *viz.* Dogri (*doi*) which are available on the project website.**

In this work, we describe word-level testsets. However, the typical usecase for transliteration is keyboard input of sentences (or at least a sequence of words). In such cases, the context would be useful to improve transliteration. A sentence-level transliteration benchmark would be useful for evaluation such contextual transliteration models. The Dakshina dataset has sentence-level transliteration testsets for 12 languages. **In a project concurrent to this work (Madhani et al., 2023), we have created sentence-level transliteration testsets for 22 Indic languages. This dataset is also available on the project website.**

In this work, we have only explored romanized to native script transliteration. However, there is a need for native script to romanized models as well for processing romanized Indic language text that is also prevalent on the web. **Subsequent to the acceptance of this work, we have also released an Indic to Roman script IndicXlit model trained on the Aksharantar corpus. This model is also available on the project website.**

## Ethics Statement

For the human annotations on the dataset, the language experts are native speakers of the languages from the Indian subcontinent. We collaborated with external agencies for the annotation task. The payment was based on their skill set and experience, determined by the external agencies, and adhered to the government's norms. The dataset is free from harmful content. The annotators were made aware of the fact that the annotations would be released publicly and the annotations contain no private information. The proposed benchmark builds upon existing datasets. These datasets and related works have been cited.

The annotations are collected on a publicly available dataset and will be released publicly for future use.

## Acknowledgements

We would like to thank EkStep Foundation for their generous grant which went into hiring human resources as well as cloud resources needed for this work. We would like to thank Vivek Seshadri and Anurag Shukla from Karya

Inc. for helping setup the data collection infrastructure on the Karya platform. We would like to thank Pratham Books for supporting the data collection in Bodo, Nepali, Urdu and Kashmiri. We would like to thank DesiCrew and Shri Samarth Krupa Language Solutions for connecting us to native speakers for annotating data. We would like to thank Anupama Sujatha for helping in coordinating the annotation work. We would also like to thank the Centre for Development of Advanced Computing, India (C-DAC) for providing access to the Param Siddhi supercomputer for training our models.